\documentclass[11pt]{article}
\usepackage{acl2015}
\usepackage{times}
\usepackage{url}
\usepackage{latexsym}
\usepackage{csquotes}
\usepackage{multirow}
\usepackage{algorithm}
\usepackage{algpseudocode}
\usepackage{amsmath}
\usepackage{amssymb}
\usepackage{float}
\usepackage[colorinlistoftodos,prependcaption,textsize=tiny]{todonotes}

\title{Low-Shot Learning for Fictional Claim Verification}

\author{Viswanath Chadalapaka \\
  {\tt vchad@usc.edu} \\
  \And
  Derek Nguyen \\
  {\tt derekngu@usc.edu} \\
  \AND
  JoonWon Choi \\
  {\tt joonwonc@usc.edu} \\
  \And 
  Shaunak Joshi \\
  {\tt shaunaks@usc.edu} \\
  \And
  Mohammad Rostami \\
  {\tt rostamim@usc.edu}}

\begin{document}

\maketitle
\begin{abstract}
    In this paper, we study the problem of claim verification in the context of claims about fictional stories in a low-shot learning setting. To this end, we generate two synthetic datasets and then develop an end-to-end pipeline and model that is tested on both benchmarks. To test the efficacy of our pipeline and the difficulty of benchmarks, we compare our models' results against human and random assignment results. Our code is available at \url{https://github.com/Derposoft/plot_hole_detection}.

\end{abstract}

\section{Introduction}
    Claim verification has become a popular research topic over the last several years as social media companies began to show a need to curb misinformation from spreading on their platforms. This endeavor has produced a large body of work focused on determining the veracity of a claim given a large body of documents containing facts (or evidence), generally pulled from Wikipedia or other domain-specific, academic corpuses. However, many practical problems outside of fake news detection which could benefit from an automated claim verification system have yet to benefit from that body of work -- specifically, those which involve scenarios where a pre-established body of facts outside of one specific work in question is not available.
    
    For example, people constantly create and consume massive amounts of entertainment through novels, TV shows, and movies. Oftentimes, the creation of these works involves substantial financial commitments, and their success amongst critics decides whether or not that investment pays off. That means that one of the most important aspects of a work is its ability to effectively capture a reader or a viewer in the story. One of the biggest detractors to that goal within these tales are the existence of logical disconnects, aka \enquote{plot holes}, which can completely destroy the credibility of a fictional world. Given what is at stake, the accurate detection of plot holes is a practical problem of interest. Although this problem is similar to fake news detection on the surface, plot hole detection is one such problem which fundamentally differs from regular claim verification tasks in that it generally occurs in a \enquote{new} setting: it concerns itself with \enquote{facts} in a fictional world, where a set of ground truth evidence statements is not well-defined. This means that regular state-of-the-art claim verification techniques cannot be applied without assuming a set of common-sense facts and risking a high rate of false positive faulty claims.
    
    For instance, in the world of \textit{Harry Potter}, magic clearly exists -- however, this directly contradicts the physical laws and common sense of reality, so assuming those facts would flag most of the work as a \enquote{plot hole}. Even if we were to use the extensive fan material and fan-made wiki pages available for \textit{Harry Potter} specifically, this would still be of very little use to us for any other new (or even unreleased) work. Given that the application of a claim verification system under these conditions must make accurate predictions for these fictional works with new, out-of-domain worlds relative to any works from the training data, we find that in general, this problem requires a solution capable of low-shot learning.
    
    To this end, we introduce a simple end-to-end pipeline for training models to solve similar problems in low-shot settings which lack labelled data, by taking advantage of knowledge graphs (KGs) and graph neural networks (GNNs). To test the efficacy of our pipeline, we define two proof-of-concept \enquote{plot hole problems} as benchmarks and implement our full pipeline for both of them to show its efficacy relative to human and random guessing benchmarks. Our results show that our pipeline generates models which successfully perform better than random guessing, and can also outperform humans annotators in some cases with statistical significance.

\section{Related Works}
    \paragraph{Claim Verification.} Current work on claim verification has focused mainly on tagging disinformation in social media spheres, and is therefore centered around fact-checking claims using Wikipedia as ground truth, e.g., the FEVER dataset \cite{fever_dataset_claimverification}. The most effective techniques for solving these problems thus far generally follow a 3-step pipeline which was first introduced along with the dataset: document retrieval, sentence-level evidence retrieval, and textual entailment. Consequently, a bulk of work has been done surrounding scientific claim verification \cite{wadden2020scifact} via search-related techniques combined with transformers such as BERT \cite{tenney2019bert}. Work has also been done regarding general fact checking in a variety of languages \cite{kazemi2021claim_bilingual}, including evidence-based fact checking \cite{alhindi2018your_factchecking_evidence}. However, document retrieval is out of the question for applying claim verification to new, fictional works. Via our proposed pipeline, our work has the potential to extend and generalize these existing solutions to low-shot settings.
    
    \paragraph{Machine Reading Comprehension.} The exploration of Machine Reading Comprehension (MRC) has resulted in the definition of several tasks and models specializing in understanding natural language questions and answering them. These tasks are sometimes claim verification-adjacent in nature. For example, recent work has described an approach to classify the right ending of a story from given right and wrong endings \cite{srinivasan-etal-2018-simple} via taking advantage of skip-thought and GloVe embeddings \cite{pennington2014glove} passed through Bi-LSTM and LSTM models \cite{huang2015bidirectional,yu2019review_lstm} to classify the correct ending of the story. This could be seen as effectively verifying the \enquote{claim} of a given ending being true. Our work contributes a solution for a subset of similar MRC task to this one by focusing on tasks in low-shot scenarios.

    \paragraph{Low-Shot Learning.} Our problem necessitates the use of models capable of low-shot classification because we need to provide accurate plot hole predictions for stories with characters, events, and rules that have not been seen before. Recent claim verification research, in addition to simply presenting results, has also aimed to assess the performance of novel methods against practical low-shot scenarios -- for example, MultiVerS \cite{wadden2022multivers}, which tests novel models trained on FEVER against out-of-domain scientific claims such as SciFact, HealthVER \cite{sarrouti2021evidence_healthver}, and CovidFact \cite{saakyan2021covid}. The results of these tests unsurprisingly shown that in-domain pretraining significantly improves claim verification performance. Since in-domain pretraining cannot be guaranteed for finding plot holes in unreleased works, in this paper, we attempt to introduce methods to improve the claim verification performance in scenarios in which there is no additional in-domain data. 
    
    
    
 \section{Problem Definition}
        As described earlier, plot hole detection is a good example of a low-shot claim verification problem that can easily be used as a proof-of-concept to demonstrate our pipeline. To be specific, we focus on the detection of two main classes of plot holes: \textit{continuity errors}, and \textit{unresolved storylines}. A \enquote{continuity error} means that a logical inconsistency is found in the story, whereas an \enquote{unresolved storyline error} means that some aspect of the story was left incomplete.
        

        More formally, we define a document $X$ made up of $n$ sentences (aka claims) $X=\{X_1, ..., X_n\}$, with $len(X)=n$. A \enquote{continuity error} occurs if there exists some sentence $X_i$ such that the remaining sentences entail the opposite meaning of $X_i$, i.e. $X \setminus X_i \models \neg X_i$. For our purposes, given a story with an unresolved error, we consider a solution to the problem to be the index $i$ of the problematic sentence $X_i$.
        
        An \enquote{unresolved storyline error} can loosely be described as the percentage of the story $p$ that has been \enquote{left out} by some arbitrary measure. If one assesses, for example, that $m$ sentences $X_{n+1},...,X_{n+m}$ are required to complete a work, then the percentage of missing sentences would be guessed as $\frac{m}{n+m}$. We admit that this problem is not as well-defined as a continuity error, but given that it is a good example of a highly practical real-world use case of our work, we pursue it regardless.

    \begin{figure*}[t!]
        \center
        \includegraphics[width=\textwidth]{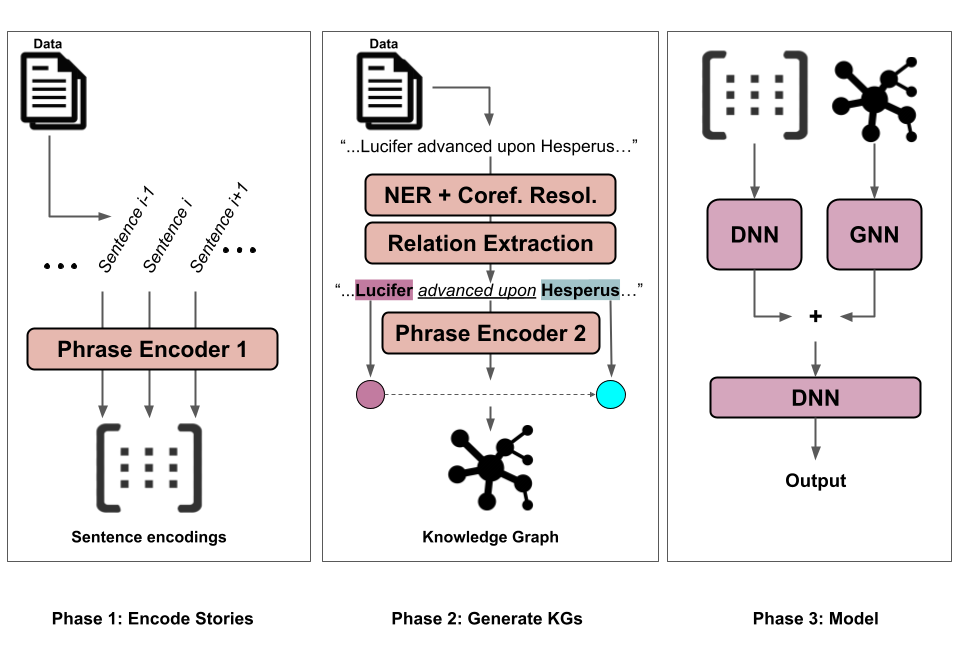}
        \caption{Visualization of our 3-phase method. First, using a phrase encoder of our choice, we create story encodings by running each sentence of a story through the encoder. Then, we generate a directed knowledge graph representing relationships between named entities in the story by performing NER, coreference resolution, and relation extraction with the help of another phrase encoder. Finally, a joint model consisting of both a GNN and a deep neural network (DNN) is used to solve the problem.}
        \label{flow_diagram}
    \end{figure*}
\section{Proposed Method}
    After defining a problem, our pipeline consists of 3 phases: two data preprocessing steps to first generate story encodings and then a knowledge graph, and then a joint graph neural network and deep neural network (DNN) model. A visualization of our pipeline is shown in Figure~\ref{flow_diagram}.

    \subsection{Dataset Preprocessing}
        Each of our datapoints is preprocessed twice: once to generate an encoding of the story, and then a second time to generate a knowledge graph (KG) of the story which is used in some of our models.
        
        \paragraph{Story Encoding.} Each story is first encoded sentence-by-sentence using a BERT model fine-tuned for producing high-fidelity 384-dimensional semantic sentence encodings called a SentenceTransformer \cite{sentence_transformer}, resulting in an encoding of size $(n, 384)$, with $n$ being the number of sentences in the given story. Since different stories have a different number of sentences in general, all stories are then zero-padded to meet the length of the longest story, resulting in a final story encoding of size $(N, 384)$, where $N$ is the length of the longest story in the entire dataset.
        
        \paragraph{Knowledge Graphs.} Since we can't use pre-established facts to determine claim verification for our tasks, KGs are a good way to formalize the information from each story that we can actually use. For our purposes, we use the Stanford CoreNLP library \cite{stanfordcorenlp_manning} after tokenization and lemmatization to extract KGs from a story in 4 steps:
        \begin{enumerate}
            \item Named entity recognition, to decide what the nodes should be in the KG.
            \item Coreference resolution, which helps build more meaningful KGs.
            \item Relation extraction (or triple extraction), in which knowledge triples (or subject-relation-object triples) are extracted on a sentence-by-sentence basis. From this point onwards, subjects and objects are considered \enquote{nodes} while the relations are considered to be \enquote{edges}.
            \item Node and edge embedding, in which nodes are one-hot encoded to represent which entity they are representative of, and a semantic encoding of edges' relation text is extracted via a SentenceTransformer.
        \end{enumerate}
        The end result of this process is a set of 3 values which comprises a single KG: node embeddings of shape $(n_{nodes}, d_n)$, edges, and edge embeddings of shape $(n_{edges}, d_e)$, where $n_{nodes}$ and $n_{edges}$ are the number of nodes and edges in the graph respectively, $d_n$ is the maximum number of entities in any given story, and $d_e$ is 384.

    \subsection{Models}
        We use a simple, transformer-based model to illustrate the efficacy of our pipeline for each of the two proposed plot hole problems. Since both models use story encodings generated by the SentenceTransformer BERT model, we refer to them as C-BERT and U-BERT for the continuity and unresolved error problems respectively. We then optionally use graph neural networks (GNNs) to process the KG input as well.
        
        \subsection{Graph Neural Networks}
            To make use of KGs, we use a type of neural network architecture called a graph neural network (GNN), which is capable of processing graph input. This is done via iterative, neighborhood-level aggregations at each node on the graph to update node features (called \textit{convolutions}). Formally, given a graph $G=(V,E)$ for $V$ the set of vertices and $E$ the set of edges, let $h_u^l$ be the $d$-dimensional node features for a node $u \in V$ at a layer $l$ in the GNN and $h_{uv}$ be $d'$-dimensional edge features corresponding to the edge between nodes $u$ and $v$. Let $N(v)$ be the set of nodes neighboring $v$. Then, in general, a GNN layer's convolution operation can be described as follows:

            \begin{equation}
            \begin{split}
            &\rho_{uv}^l = \rho (h_u^l, h_v^l, h_{uv}; \Theta_\rho) \\&
            m_u^l = \zeta (\{\rho_{uv}^l | v \in N(u)\}; \Theta_\zeta) \\&
            h_u^{l+1} = \phi(m_u^l, h_u^l; \Theta_\phi) 
            \end{split}
            \end{equation}
            
            where $\rho$, $\zeta$, and $\phi$ represent functions with learnable parameters whose definitions differ depending on the type of GNN used \cite{genconv_conv}. 

            In this paper, the GNN we use we use to process our KGs is a more expressive variant of the popular graph attention network \cite{gat_velickovic2017graph} called the graph attention network v2 (GATv2) \cite{gatv2}. GATv2 convolutions can be described by the following $\rho$, $\zeta$, and $\phi$:

            \begin{equation}
            \begin{split}
            &\rho = a^T LeakyReLU (W \cdot [h_u^l || h_v^l] )\\&
            \zeta = \sigma (\sum_{\rho^l_{uv} \in S} softmax_v(\rho^l_{uv}) \cdot Wh_j)\\&
            \phi = m_u^l
            \end{split}
            \end{equation}

            where $\Theta_\rho=\{a, W\}$ for $W \in \mathbb{R}^d, a \in \mathbb{R}^{2d}$, $\Theta_\zeta = \{W\}$ (the same weights as are used in $\Theta_\rho$), and $\Theta_\phi = \emptyset$.
        
        \subsection{C-BERT and U-BERT}
            Both of our models take a sequence of encoded sentences as input, and process them by first taking them to an embedding dimension and a series of transformer encoder layers. Since the use of KGs in low-shot scenarios are a target of our ablation study, each model can be configured either to use or not use GNNs and KG inputs.
            
            From there, C-BERT returns a set of probabilities over all of the sentences which represent each sentence's likelihood of containing a continuity error. This is done via a feedforward network (FFN) and a softmax layer. If KGs are being utilized, then the output of the GNN processing the KG is concatenated to each sentence encoding before applying the FFN.
            
            U-BERT returns a single real number representing the portion of sentences that it predicts were left off the end of the given story. This is done by applying transformer decoders to take the input sequence to an output sequence of length 1, and then similarly applying a FFN followed by a sigmoid. If KGs are utilized for this model, the GNN output is similarly concatenated to the result of the transformer decoder before applying the FFN.

\section{Experimental Setup}
    \subsection{Dataset}
        For our experiments, the generation process described under our methods was applied on 2000 fictional stories pulled from the r/WritingPrompts and r/stories subreddits and sourced from Kaggle, all of which had at least 1000 upvotes and at least 200 words. This resulted in two separate labelled datasets of 2000 samples each -- one for each type of plot hole. Within each dataset, 1000 samples were used for training, and the other 1000 for testing.
        
    \subsection{Synthetic Data Generation}
        Since we are primarily interested in scenarios which lack labelled datasets and we are using supervised learning, synthetic data generation is an important step for our pipeline. To generate a synthetic dataset, we deliberately inject plots holes (i.e., false claims) into documents for each of the two problems, which generates two synthetic datasets. This method of creating a dataset is similar to the way that the FEVER dataset is generated. 
        
        Formally, for a sentence $s$, let $v(s)$ be a function that returns the first verb of $s$, and for some word $w$, let $A(w)$ be a function that returns a list of possible antonyms for $w$ ordered by relevance. For a verb $v$, we also define $C(v)$ as the set of conjugations of $v$ (e.g., $C(``be") = \{are, is, was ...\}$).
        
        A \textit{continuity error} is injected into a story $X$ by uniformly choosing a random sentence $s$ from it and negating its meaning, as presented in   Algorithm~\ref{continuity_error_injection_alg}. The label is then the index of the negated sentence. \textit{Unresolved storylines} can similarly be injected by removing between 0\% and 10\% of the endings of stories. The label is then the percentage of the story that was removed. For simplicity, negative examples (i.e. unaltered stories) are not included in our final dataset.
        
        Our benchmarks for each dataset show that humans are expected to score ~$0.5$ F1 on the continuity error dataset and ~$2.51e-3$ MSE on the unresolved error dataset (on $n=10$ stories).

        \begin{algorithm}[t!]
            \caption{Continuity error injection method}
            \label{continuity_error_injection_alg}
            \begin{algorithmic}
                \Require document $X$
                \State $y \gets unif\{1, len(X)\}$
                \State $s \gets X_y$
                \State $v \gets v(s)$
                \If{$v \in C(``be")$ \textbf{or} $\nexists A(V)$}
                    \State $v \gets v+``not"$
                \Else{$\exists A(v)$}
                    \State $v \gets A(v)[0]$
                \EndIf
                \State $s \gets s$ with updated $v$
                \State $X_y \gets s$
                \State \Return $X, y$
            \end{algorithmic}
        \end{algorithm}
        
        \begin{algorithm}[t!]
            \caption{Unresolved error injection method}
            \label{unresolved_error_injection_alg}
            \begin{algorithmic}
                \Require document $X$
                \State $y \gets unif\{\lfloor 0.9*len(X) \rfloor, len(X)\}$
                \State $X \gets X_{1:y-1}$
                \State $p \gets \frac{y}{n}$
                \State \Return $X, p$
            \end{algorithmic}
        \end{algorithm}
    
    \subsection{Baseline Methods}
        We provide human annotator results on our synthetic dataset to be used as a benchmark for them. Since this is a newly-introduced dataset, we also give the expected value of results obtained by guessing, to show statistical significance of our models even in cases where they do not outperform the human benchmark.
    
    \subsection{Evaluation Metrics}
        The continuity error dataset is scored using F1 score (as is standard for regular claim verification), and the unresolved error dataset is scored using the MSE between the actual and predicted percentage of the removed story. A higher F1 score is considered better, and likewise a lower MSE loss is considered better. To decide which model performed the best, models are trained using different seeds to obtain a distribution of metrics and statistical significance is shown via 1-sample or 2-sample T-tests with $p<0.01$. The expected values of the guessing model assume that a random sentence is guessed for continuity errors, and that 0.05 is guessed every single time (as it is exactly in between the 0 and 0.1 used to generate the data) for unresolved errors.

\section{Results and Conclusion}
    \begin{center}
    \label{results}
    \begin{table}[ht]
        \centering
        \begin{tabular}{ c|c }
            Model & F1 \\
            \hline
            Guessing & 0.026 \\ 
            Human & 0.500 \\ 
            C-BERT & 0.182 $\pm$ 0.004 \\ 
            C-BERT+GAT & \textbf{0.195 $\pm$ 0.009} \\
        \end{tabular}
        \caption{\textbf{Continuity Error Model Results.}}
        \label{continuity_results}
    \end{table}
    \begin{table}[ht]
        \centering
        \begin{tabular}{ c|c }
            Model & MSE \\
            \hline
            Guessing & 1.37e-2 \\ 
            Human & 2.51e-3 \\
            U-BERT & \textbf{4.69e-4 $\pm$ 9.08e-6} \\ 
            U-BERT+GAT & 8.71e-3 $\pm$ 1.49e-4 \\ 
        \end{tabular}
        \caption{\textbf{Unresolved Error Model Results.} Bolded values indicate best statistically significant results (excluding human benchmarks). Model training results are accompanied by a 95\% confidence interval over $n=5$ runs.}
        \label{unresolved_results}
    \end{table}
    \end{center}
    
    For the continuity error problem, our C-BERT model does not outperform a human annotator. Despite that, the C-BERT+GAT results show that the use of a KG as input to our model results in a statistically significant lift in performance. These results form a strong basis for employing similar techniques with KGs for performance gains in adjacent MRC problems.
    
    For the unresolved error problem, our U-BERT model outperforms both random guessing and human annotators by a statistically significant margin. On the other hand, our U-BERT+GAT results -- which underperform both the U-BERT and human benchmark -- suggest that incorporating KGs does not necessarily result in higher performance for every class of low-shot fictional claim verification problem.

    In conclusion, our results show that both of our proposed models outperform random guessing with statistical significance, and in some cases can also outperform human annotators, thus establishing a reasonable first baseline on our benchmark problems. This proves that our pipeline has the potential to produce high-performing models for low-shot fictional claim verification problems without the need for labelled data. We hope that our results help to advance the adoption of claim verification solutions outside of social media, especially given its potential impact to the entertainment industry.

\bibliographystyle{acl}
\bibliography{acl2015.bib}

\end{document}